\documentclass[nohyperref]{article}

\usepackage{microtype}
\usepackage{graphicx}
\usepackage{subfigure}
\usepackage{booktabs} 
\usepackage[section]{placeins}
\usepackage{makecell}

\usepackage{hyperref}

\usepackage[accepted]{icml2022}

\usepackage{amsmath}
\usepackage{amssymb}
\usepackage{mathtools}
\usepackage{amsthm}

\usepackage[capitalize,noabbrev]{cleveref}

\theoremstyle{plain}

\theoremstyle{definition}

\theoremstyle{remark}

\usepackage[textsize=tiny]{todonotes}

\icmltitlerunning{Speeding up Resnet Architecture with Layers Targeted Low Rank Decomposition}

\begin{document}

\twocolumn[
\icmltitle{ Speeding up Resnet Architecture with \\ Layers Targeted Low Rank Decomposition }



\icmlsetsymbol{equal}{*}

\begin{icmlauthorlist}
\icmlauthor{Walid Ahmed}{comp}
\icmlauthor{Habib Hajimolahoseini}{comp}
\icmlauthor{Austin Wen}{comp}
\icmlauthor{ Yang Liu}{comp}

\end{icmlauthorlist}

\icmlaffiliation{comp}{Ascend Team, Toronto Research Center,\\ Huawei Technologies, Toronto, Canada}

\icmlcorrespondingauthor{Walid Ahmed}{walid.ahmed1@huawei.com}

\icmlkeywords{Machine Learning, ICML}

\vskip 0.3in
]



\printAffiliationsAndNotice{}  

\begin{abstract}

Compression of a neural network can help in speeding up both the training and the inference of the network. In this research, we study applying  compression using low rank decomposition  on network layers.  Our research demonstrates that to acquire a speed up, the compression methodology should be aware of the underlying hardware as  analysis should be done to choose which  layers to compress.  
The advantage of our approach is demonstrated via a case study of compressing ResNet50 and training on full ImageNet-ILSVRC2012. We tested on two different hardware  systems
Nvidia V100  and Huawei Ascend910. 
With hardware targeted compression,  results on Ascend910  showed  5.36\% training speedup and  15.79\% inference speed  on Ascend310 with only 1\% drop in accuracy  compared to the original uncompressed model.
 
\end{abstract}

\section{Introduction} 
\label{intro}


With the past researches and development in the field of deep learning, the size of the models have grown bigger and bigger, the scale of the trainable parameters has reached billions or even higher.
Therefore, training such large models had become a critical problem for it's usually time and memory consuming \citep{dean2012large, hajimolahoseini2023methods, li2021short}. 
The problem is even more severe on smart phones and edge devices, memory usage and computational costs are often non-negligible, considering the limited computational resources and battery life e.g. on portable medical devices \citep{hajimolahoseini2022long, hajimolahoseini2012extended, hajimolahoseini2016robust, hajimolahoseini2018inflection}.
However, it’s possible to compress a model several times smaller, while also preserving the accuracy close to the originals, due to the high redundancy in most model structures \cite{chen2019drop,hajimolahoseini2023training, hajimolahoseini2022strategies, hajimolahoseini2019deep}.

Current techniques for deep learning acceleration can mainly be categorized into several groups: Low Rank Decomposition (LRD), pruning, quantization and knowledge distillation (KD) \citep{cheng2017survey, hajimolahoseini2018ecg}. 
In deep learning models, the objective of compression is to optimize the computational complexity and memory consumption without changing the network architecture, except for KD, the structure of the model could be different before and after the compression \citep{hinton2015distilling, rashid2021mate}.


In pruning, compression is achieved by deleting the nodes that contribute less for the final performance of the model, as the result, the sparsity of the model is increased \cite{luo2018thinet}. 
The technique can be used on different network components such as kernels, filters, kernels, or weights of the layers \cite{zhang2018systematic, zhuang2018discrimination, mao2017exploring}. 
A heuristic ranking measure guides the implementation of pruning, which is usually determined based on experiments.  
The biggest drawback of pruning as a compression technique is that it could take an extended time to pruning and for fine-tuning to converge, this could cause training overhead \citep{cheng2017survey}. 

Quantization is the technique of replacing floating-point numbers of a network's weight to lower bit width numbers. Different precision and ways to apply quantization can be adopted.
For example fixed-point quantization, the weights are recast into lower precision e.g. binary or 8-bits \citep{prato2019fully, bie2019simplified}. 

A teacher-student framework is applied for Knowledge Distillation (KD), this enables the transfer of the knowledge from the teacher network into a smaller and more efficient network (student). The process includes optimization of two losses between the label, student model's output and teacher model's output. \citep{hinton2015distilling, rashid2021mate}.
However, lack of mathematical support made these methods limited when facing high compression ratios situation.

Compared with methods mentioned above, Low Rank Decomposition (LRD) utilize a tensor decomposition algorithm to decompose weight parameters into smaller matrices. e.g. Singular Value Decomposition (SVD) \cite{hajimolahoseinicompressing, van1987matrix}.
This algorithm replaces each fully-connected layer with two fully-connected layers of different weight matrices, and these two matrices are derived from the original one using SVD.
For convolutional layers Tucker is used  to decompose them into multiple components \cite{de2000multilinear, de2000best}.

LRD could provide several benefits, including a strong mathematical support and the knowledge distillation gained from original model, but one side-effect is applying tensor decomposition will generate high number of new layers,this will prevent the method of being considered a training/inference acceleration technique in terms of fps.

In this research, we are using  lrd with  tensor decomposition algorithms to compress a neural network. Our aim is to have a faster training and as well as faster inference. The paper is organized as follows. After this introductory  section, a section will follow that present our proposed method and  outline how can we assure that applying lrd we will have a speedup. Section  \ref{caseStudy} offer a complete case study where we apply our method on the benchmark ResNet50 \cite{https://doi.org/10.48550/arxiv.1512.03385}. Finally section \ref{conclusion}    
conclude the research.

\section{Method}

In our research we claim that based on the underling hardware, not all the neural network layers are good candidates for compression. Some layers are "compressible friendly" while other are not.
In our method we  keep a list of which  layers to compress, other layers that are not in the list will not be compressed and will be kept as is. While there are practical experiments  that can be applied to find the optimum list of layers to compress, this might not be applicable with the large number of layers. As an alternative, we introduce of the concept of a compression mode. where each mode represent a group of layers. These modes are tested to find out the best mode. With LRD, We found that it is more challenging to gain the speedup for the model than to  keep the accuracy drop within a certain limit. Proper learning rate schedule can be tuned to improve the accuracy but the speedup should be guaranteed first. Choice of targeted compression ratio should be reasonable, in our method we targeted   a 2x compression  with an accuracy drop within a 1\% from the accuracy of the original uncompressed model.

\begin{figure}[h]
\vskip 0.2in
\begin{center}
\centerline{\includegraphics[width=\columnwidth]{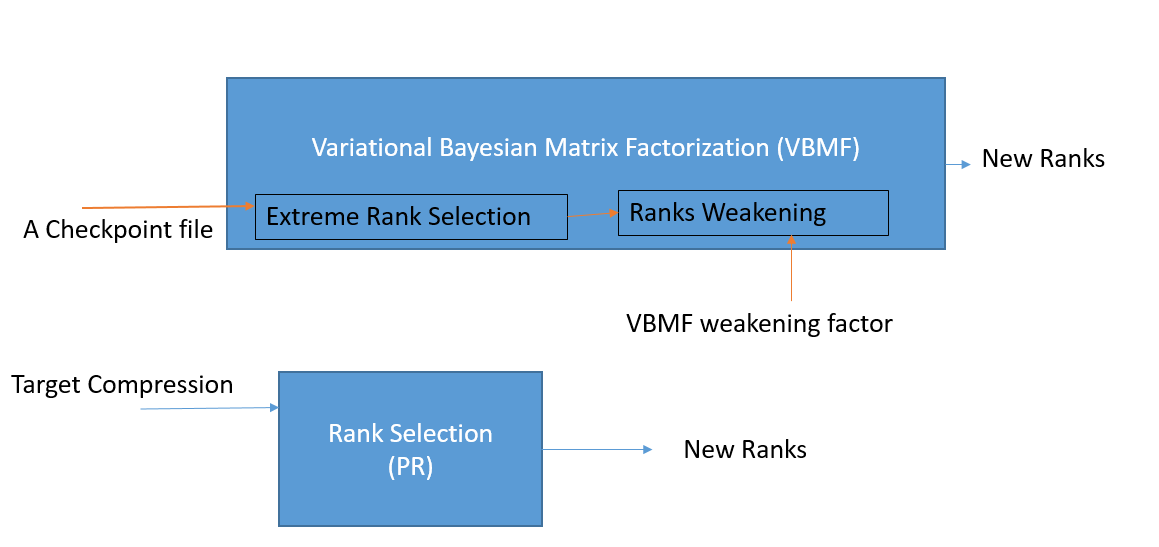}}
\caption{Rank Selection.}
\label{rank-selectionPipeLine}
\end{center}
\vskip -0.2in
\end{figure}

Model compression using LRD by tensor decomposition  starts with training  the original  architecture  for a total  of N$_1$ epoch. After training is done, we will have a checkpoint with the updated weights. Following that, based on  our selection of which layers to compress, the next step will be to apply the rank selection. Rank selection can be done in an intuitive way of parameters reduction (PR)  where the ranks will be calculated based on the targeted  compression ratio. Parameters reduction  does not use the weights in trained checkpoint.
The other method to apply  rank selection  is using the  Variational Bayesian Matrix Factorization  method \cite{https://doi.org/10.48550/arxiv.1801.05243} which aims more into decreasing the reconstruction error by considering the weights in the trained checkpoint of the original model. Using VBMF weakening factor \cite{DBLP:journals/corr/abs-1803-08995}, we can have some control on the final compression ratio. Rank selection pipeline is shown in figure \ref{rank-selectionPipeLine}.

\begin{figure}[ht]
\vskip 0.2in
\begin{center}
\centerline{\includegraphics[width=\columnwidth]{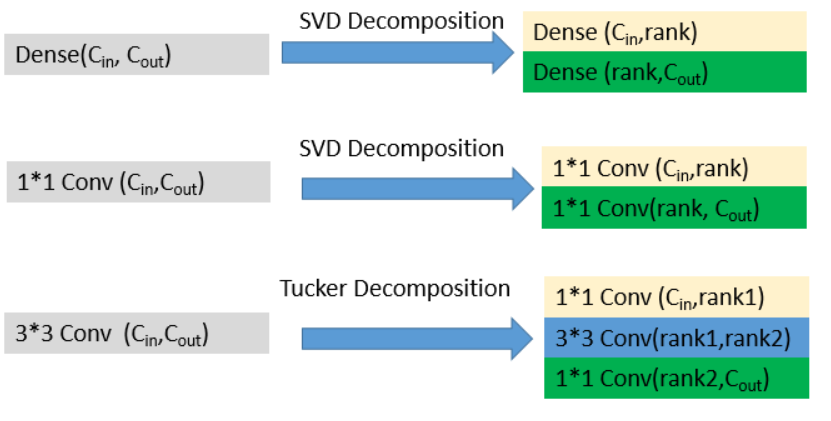}}
\caption{Layer Decomposition.}
\label{Sample_Layer_Decomposition}
\end{center}
\vskip -0.2in
\end{figure}

\begin{figure}[ht]
\vskip 0.2in
\begin{center}
\centerline{\includegraphics[width=\columnwidth]{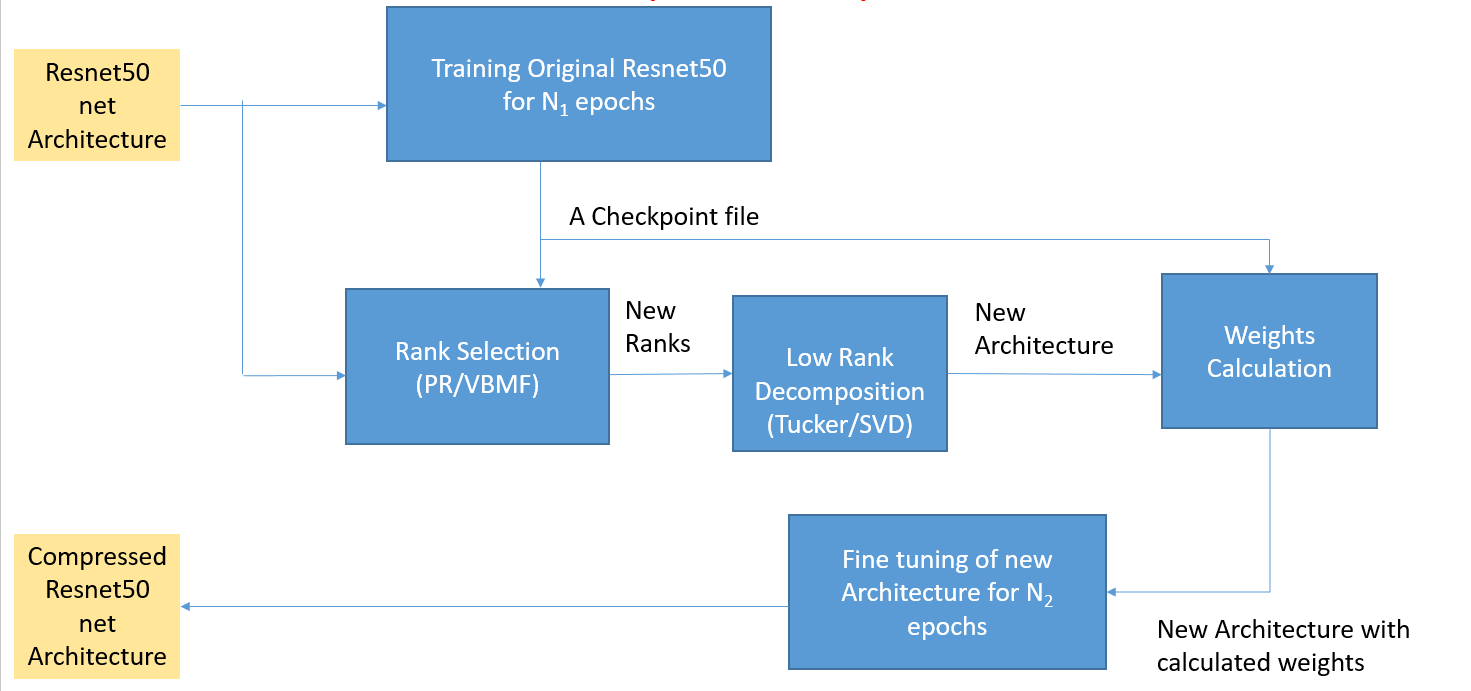}}
\caption{Compression Framework Pipeline.}
\label{Compression_Framework_Pipeline}
\end{center}
\vskip -0.2in
\end{figure}

Based on the calculated  ranks, we  apply  low rank decomposition on the weights tensors of our  selected layers in the network. This will result in a new architecture which will have more layers but the total number of parameters will decrease which will also results in a decline in the number of flops. A 1*1 conv  layer is decomposed into two successive 1*1 conv layers, a 3*3 conv layers is  decomposed into 3 layers: a 1*1 conv layer followed by a 3*3 conv layer followed by a 1*1 conv layer,  a dense layer is decomposed into   two dense layers. The new architecture is our compressed model.  Figure \ref{Sample_Layer_Decomposition} illustrates the layers decomposition. 
The weights of the decomposed layers will be calculated aiming to have the least reconstruction error. After the new weights are applied, one of the measurements that can be used is   the accuracy of the compressed model at this time. After that the compressed model is fine tuned by training for $N_2$ epochs. The pipeline for compressing Resnet50 architecture is  shown in figure  \ref{Compression_Framework_Pipeline}.

Our proposed method  also includes 2 more components. 
\begin{itemize}
  \item Final Dense Layer Compression Rate: As the final accuracy of the compressed model in convolution neural network will relay on the the final dense layer(s). We can benefit form having a lower  compression rate  for them    than the other convolution layers. 
  \item Ranks Quantization: The  training and inference speed  on a certain  hardware might benefit from quantization of ranks of layers  to be multiple of a certain number which we call Ranks Quantization.  After the ranks are calculated, there values will be quantized as favoured by the hardware.
\end{itemize}

\section{Case Study}
\label{caseStudy}

\subsection{Applying the LRD pipeline on Resnet50}

\begin{figure}[ht]
\vskip 0.2in
\begin{center}
\centerline{\includegraphics[width=\columnwidth]{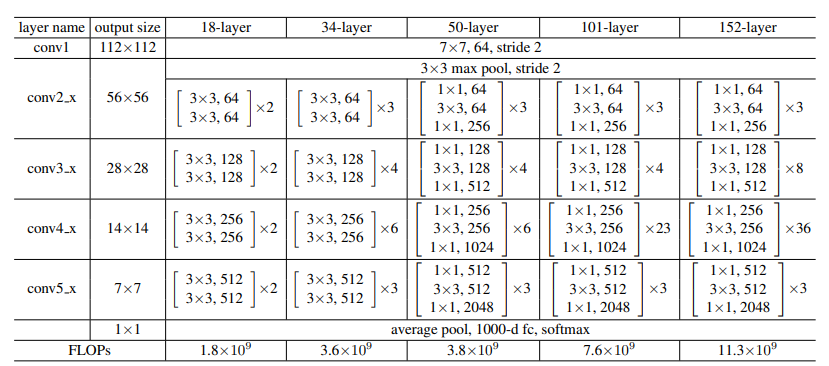}}
\caption{Resnet Architectures}
\label{Resnet50}
\end{center}
\vskip -0.2in
\end{figure}

Our case study will be lrd compression of  Resnet50\cite{https://doi.org/10.48550/arxiv.1512.03385}  trained on the on full ImageNet-ILSVRC201 \cite{5206848,awad2023improving}. A description of the different architecture of ResNet is shown in figure \ref{Resnet50}. Each ResNet architecture has 4 components namely conv2\_x,conv3\_x,conv4\_x and conv5\_x, we refer to these components as blocks 1,2,3 and 4  respectively.

We start with training the original  model. The accuracy and the speed  of this uncompressed model will be our benchmark. An intermediate checkpoint from the training will be saved and used as an input in  the compression  pipeline which we showed in figure \ref{Compression_Framework_Pipeline}. The compression can be done with different modes,these modes are listed in table \ref{Compression_Modes}. While the vanilla mode includes compression of all the layers in the 4 blocks, other modes had different combinations of layers.  

Training for   ResNet50 was done for 90 epochs, checkpoint from epoch from epoch 45 was saved to be used later for model compression. Fine tuning of the compressed model was done for 45 epochs. The fine tuning of the compressed model required a concern for the learning rate schedule. As the decomposed weights originated from a checkpoint, typically we  should start from a lower learning rate.

\begin{table}[t!]
\caption{Compression Modes}
\label{Compression_Modes}
\vskip 0.05in
\begin{center}
\begin{small}
\begin{sc}
\begin{tabular}{l|p{30mm}}
\toprule
Mode &  Description \\
\midrule 
Vanilla   & All conv layers in 4 blocks with Final Dense Layer. \\
Mode1  & Vanilla Compression  excluding 1*1 conv layers. \\
Mode2  & Mode1 with adding   the 1*1 conv layers excluding downsample layers of blocks 3,4.\\
Mode3  & Mode2 but excluding 1*1 conv layers of block4. \\
Mode4  & Vanilla Compression  without any downsample layers. \\
Mode5  & All 1*1 conv layers in 4 blocks  with Final Dense Layer. \\
\bottomrule
\end{tabular}
\end{sc}
\end{small}
\end{center}
\vskip -0.1in
\end{table}
\subsection{Training Uncompressed ResNet50}

Training  of  the original Resnet50 using the parameters in table \ref{Original_Model_Training_Parameter},  we were  able to reach a validation accuracy of \%76.95. Using the profiler the training step-time  was found to be 114.42ms.

\begin{table}
\caption{Original Model Training Parameters}
\label{Original_Model_Training_Parameter}
\vskip 0.05in
\begin{center}
\begin{small}
\begin{sc}
\begin{tabular}{cc}
\toprule
PARAMETER &  VALUE \\
\midrule
batch size  & 256 \\
steps per epoch  & 5004 \\
momentum  & 0.9\\
epoch size  & 90 \\
label smooth factor & 0.1 \\
lr decay mode & Linear \\
lr init & 0 \\
lr max & 0.1 \\
lr end & 0 \\
\bottomrule
\end{tabular}
\end{sc}
\end{small}
\end{center}
\vskip -0.1in
\end{table}

\subsection{Applying Low Rank Decomposition on  Resnet50}

\begin{figure}[ht]
\vskip 0.2in
\begin{center}
\centerline{\includegraphics[width=\columnwidth]{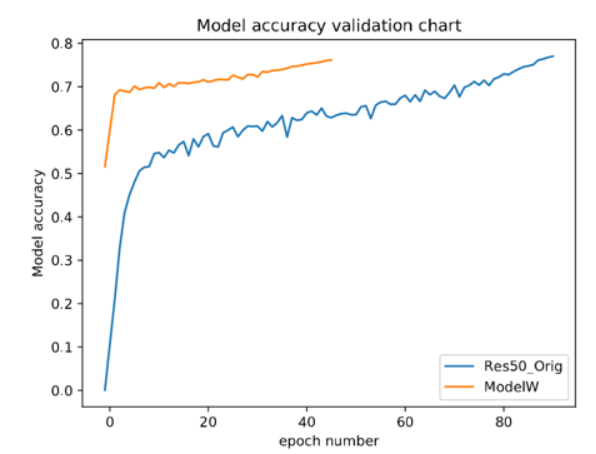}}
\caption{Training Curve original vs Vanilla Compressed}
\label{Training-Curve_original_vs_Vanilla_Compressed}
\end{center}
\vskip -0.2in
\end{figure}

Our first attempt to speed up the model on Ascend910  was to apply tensor decomposition using the mode Vanilla Compression by compression of all the convolution layers in the four blocks  and the final dense layer in Resnet50.
The model was successfully compressed, the number of total parameters decreased as well as the number of flops. After fine tuning The drop in accuracy  for the compressed model was in the acceptable range. However, there had been no speedup. On the contrary, there had been a slowdown of \%18.4 with a final accuracy of 76.1\%.   Applying the same technique on Nvidia V100 resulted on a speedup of 6.8\%   with a final accuracy of  76.6\%.  Results are summarized in table \ref{origina_vanilla}.
The training curve of both the original model vs the vanilla compressed model(ModelW) on Ascend910 is shown in figure \ref{Training-Curve_original_vs_Vanilla_Compressed}.

\begin{table}[ht]
\caption{Compression of Resnet50 on Ascend910 }
\label{origina_vanilla}
\vskip 0.05in
\begin{center}
\begin{small}
\begin{sc}
\begin{tabular}{llllll}
\toprule
Model&  Parameter & Compress & FLOPs & Step Time & ACC\\
\midrule
Resnet50  & 25M &  n/a &7.8M & 114ms &  76.9\%\\
ModelW  & 13M &  2x &5.3M & 135ms &76.1\%\\
Model\_LRD & 13M &  2x &5.3M & 108ms &75.9\%\\
\bottomrule
\end{tabular}
\end{sc}
\end{small}
\end{center}
\vskip -0.1in
\end{table}

As Vanilla compression  was not able to produce speed up on  Ascend910, the next step was to experiment with the different modes of compression listed in table \ref{Compression_Modes}. We tested different compression modes and the best we found was Mode3. With this mode and other proper settings shown in table \ref{Compression_Setting_Parameters},  we were able to reach a speed up 0f 5.6\% in training on Ascend910 and 15.79\% inference speed up (measured on Ascend310). We  used a checkpoint from epoch 45 of the original model. The accuracy of the checkpoint was 64\% before decomposition, after the decomposition and before any fine tuning of the compressed model the accuracy  was 17.24\%.
The results of our best compressed model(Model\_LRD) on Ascend910 are  shown in table   \ref{origina_vanilla}.

\begin{table}
\caption{Compression Setting Parameters}
\label{Compression_Setting_Parameters}
\vskip 0.05in
\begin{center}
\begin{small}
\begin{sc}
\begin{tabular}{cc}
\toprule
PARAMETER &  VALUE \\
\midrule
Rank Selection & PR \\
Compression Mode  & Mode3 \\
Final Layer Compression & 1.3x \\
Targeted Compression    & 3x \\
Ranks Quantization & 32 \\
lr max & 0.02 \\
batch size  & 256 \\
epoch size  & 45 \\

\bottomrule
\end{tabular}
\end{sc}
\end{small}
\end{center}
\vskip -0.1in
\end{table}

\section{Conclusion}
\label{conclusion}

In this research, a framework that uses Low rank decomposition was applied to compress the Resnet50 architecture on two different hardware systems. From the results we can conclude that decreasing the number of total parameters or the total number of flops might not be always enough to speed up the model. Based on different hardware and the low level actual implementation  of  the operators in each layer,compression benefits might vary. Picking the best  set of layers to compress is not a trivial task, for that we introduced the idea of  different compression modes that should  be examined aiming to find the optimum mode.

\clearpage

\bibliography{references}
\bibliographystyle{icml2022}


\end{document}